\documentclass[a4paper, 11pt]{article} 

\usepackage[protrusion=true,expansion=true]{microtype} 
\usepackage{graphicx} 
\usepackage{wrapfig} 
\usepackage{graphicx}
\usepackage{amssymb}
\usepackage{amsmath}
\usepackage{algorithm}
\usepackage{algpseudocode}
\DeclareMathOperator*{\argmax}{argmax}
\usepackage{csquotes}
\usepackage[onehalfspacing]{setspace}
\usepackage{mathtools}
\usepackage{subcaption}

\newcommand*{\permcomb}[4][0mu]{{{}^{#3}\mkern#1#2_{#4}}}

\newcommand*{\comb}[1][-1mu]{\permcomb[#1]{C}}

\usepackage{times}
\usepackage{mathpazo} 
\usepackage[T1]{fontenc} 
\makeatletter
\renewcommand\@biblabel[1]{\textbf{#1.}} 
\renewcommand{\@listI}{\itemsep=0pt} 

\renewcommand{\maketitle}{ 
\begin{flushright} 
{\LARGE\@title} 

\vspace{50pt} 

{\large\@author} 
\\\@date 

\vspace{40pt} 
\end{flushright}
}

\title{\textbf{A Brandom-ian view of \mbox{Reinforcement Learning} towards strong-AI} 
}
\author{\textsc{Atrisha Sarkar} 
\\ \textit{David R Cheriton School of Computer Science}
\\\textit{University of Waterloo}} 

\date{\today} 

\begin{document}

\maketitle 



\begin{abstract}
The analytic philosophy of Robert Brandom, based on the ideas of pragmatism, paints a picture of sapience, through inferentialism. In this paper, we present a theory, that utilizes essential elements of Brandom's philosophy, towards the objective of achieving strong-AI. We do this by connecting the constitutive elements of reinforcement learning and the Game Of Giving and Asking For Reasons. Further, following Brandom's prescriptive thoughts, we restructure the popular reinforcement learning algorithm A3C, and show that RL algorithms can be tuned towards the objective of strong-AI.  
\end{abstract}

\vspace{30pt} 


\section{Introduction}

The analytic philosophy of Robert Brandom situates itself with the \textit{demarcation} question: \textit{what does it mean to be us?} In the process, Brandom presents a layer-cake picture of sapience, thus categorizing agents into one of three categories: \textit{simple performers, rational being, and logical beings,} in increasing order of their sapience. Brandom's approach to evaluate a being's sapience is through the Game Of Giving and Asking For Reasons (GOGAR). Brandom claims that this game is a prototypical representation of a social practice, and it is possible to evaluate a being's sapience just by their nature of participation within that game. It is also possible to draw parallels between Brandom's treatment of sapience, and John Searle's distinction of \textit{weak} and \textit{strong} AI; where the highest level of sapience in Brandom's philosophy (logical beings) is strong-AI \cite{searle1980minds}. In relation to artificial intelligence, this presents a practical approach to re-imagine models with the purpose of advancing them towards strong-AI.

In this paper, we draw the theoretical links between one such class of models i.e. reinforcement learning, and Brandom's philosophy of sapience. We do this by connecting the constitutive elements of reinforcement learning and Markov Decision Processes (MDPs), to Brandom's ideas of \textit{inferentialism}. Further, with the objective of exploring a trajectory towards strong-AI, we restructure the popular reinforcement learning Asynchronous Advantage Actor-Critic (A3C), in accordance with Brandom's philosophy. Finally, looking through the lens of Brandom's philosophy, we show that it is in principle possible to re-imagine reinforcement learning algorithms towards that objective.   

\section{Reinforcement Learning}

Reinforcement learning (RL) is loosely inspired by the psychology of rewards and punishment. This is based on the neurological evidence of dopamine release in a mammal brain, which functions to shape reward driven behavior. RL is one of the popular machine learning techniques along with supervised and unsupervised learning, where the focus is on goal directed learning. Mathematically, reinforcement learning is modeled using a Markov Decision Process (MDP) \cite{sutton1998reinforcement}. MDP is a discrete \footnote{most principles of MDP translate directly to a continuous version of MDP} state transition model which consists of:
\begin{itemize}
\item States ($S$): The set of states the agent can be in. This helps the agent to form a representation of the environment based on its perception. For e.g. sensor readings of a robot
\item Actions ($A$): The set of actions the agent can take.
\item Transition function ($T(S'|S,A)$): This is the environment model, and is expressed as a probability distribution over states that the agent can land into ($S'$), if it takes the action $A$ in state $S$,   
\item Reward function ($R(S,A,S')$): The real valued reward function that the agent receives as a result of taking an action $A$ in state $S$. Popular notation represents reward function as being dependent on the \textit{next} state ($S'$) as well as on the current state and action ($S,A$).  
\item Discount factor ($\gamma$): This controls the preference the agent gives to the immediate rewards as compared to rewards in the future.
\end{itemize}

\begin{figure}
\center{
  \includegraphics[width=0.75\linewidth]{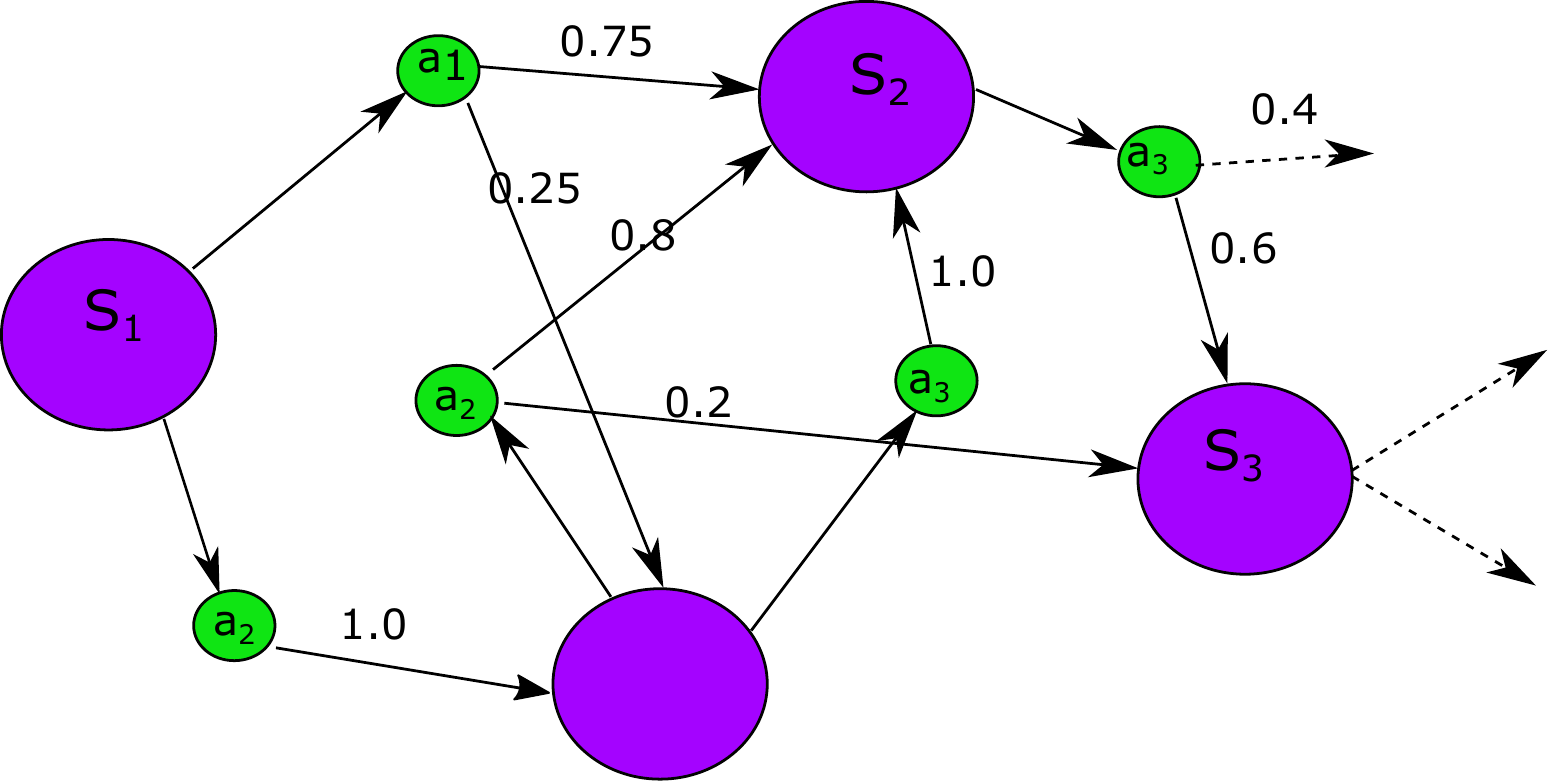}
  \caption{A Markov Decision Process with state transition probabilities, and reward values shown within braces.}
  \label{fig:mdp}
  }
\end{figure}

Figure \ref{fig:mdp} shows an MDP with the state transition probabilities, and the corresponding reward values that the agent receive for each state-action pair. RL being a goal-oriented learning mechanism, the agent progressively learns a sequence of actions (\textit{policy)} to take in order achieve a pre-defined objective. Often, the learning process is framed in a manner such that the sequence of actions that maximizes the rewards gathered along the way (\textit{optimal policy)}, results in achieving that objective.

To facilitate the learning process, the agent maintain one or more of the following functions:

\begin{itemize}
\item a \textit{policy} function $\pi: S \rightarrow A$ that maps states to actions the agent can take in that state.
\item a \textit{state-value} function $V: S \rightarrow \mathbb{R}$ that represents the goodness of the state \footnote{A note on the notation: Capitalized $V$ and $Q$ is used for tabular representation, whereas $v$ and $q$ is used for functional representation.}. It is the expected discounted sum of future rewards the agent can receive from that state onward.
\[
v_\pi(s) \doteq \mathbb{E}_\pi[G_t | S_{t}=s] = \mathbb{E}_\pi \left[\sum^{\infty}_{k=0} \gamma^{k}R_{t+k+1} | S_{t}=s \right]
\]

where $\mathbb{E}_\pi[\cdot]$ is the expected value of a random variable, given that the agent follows the policy $\pi$, $t$ is the time step, and $G_t$ is the expected sum of return from state $s$. 
\item an \textit{action-value} function $Q: S \times A \rightarrow \mathbb{R}$ that is similar to the state-value function, but represents the goodness of taking an action, given a certain state.
\[
q_\pi(s,a) \doteq \mathbb{E}_\pi[G_t | S_{t}=s,A_{t}=a] = \mathbb{E}_\pi \left[\sum^{\infty}_{k=0} \gamma^{k}R_{t+k+1} | S_{t}=s,A_{t}=a \right]
\]
\end{itemize}

\subsection{Algorithms in Reinforcement Learning}
\label{sec:algos}

\begin{figure}[htbp]
\center{
  \includegraphics[width=0.4\linewidth]{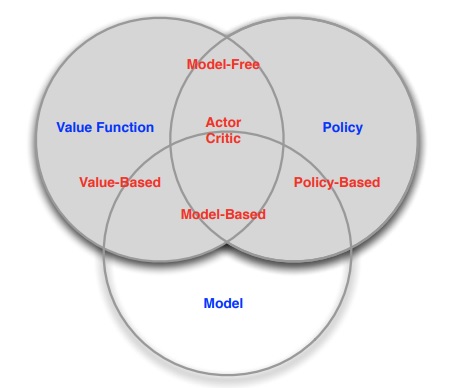}
  \caption{Categorization of various algorithms for reinforcement learning.}
  \label{fig:rl_algorithms}
  }
\end{figure}

Figure \ref{fig:rl_algorithms} shows the main categorizes that popular reinforcement learning algorithms \footnote{http://www0.cs.ucl.ac.uk/staff/d.silver/web/Teaching.html}. This section gives a short description of each category, and subsequently elaborates the algorithms that are relevant for the development of the philosophical connections in the later sections.

\textit{Value Function} based algorithms like the Q-learning \cite{watkins1992q} and SARSA \cite{sutton1996generalization} maintains an explicit representation of the $q$ or the $v$ functions, and use them to find an optimal policy of behavior ($\pi^*$). On the other hand direct \textit{policy based} algorithms like REINFORCE \cite{williams1988use}, finds an optimal policy of behavior without maintaining an explicit representation of the $q$ or $v$ functions. Actor-critic \cite{sutton1998reinforcement} are a class of algorithms that combine the value function based and direct policy based methods into a single algorithm.

\textit{Model-based} and \textit{model-free} are another categorization of RL algorithms. Model-based methods maintains and learns an explicit representation of the transition function ($T$), and uses that to calculate the value functions, subsequently deriving the optimal policy. Whereas, model-free approaches do not maintain an explicit representation of the transition function.

\subsubsection*{Q-learning}
Q-learning is model-free algorithm that iteratively learns, as the name suggests, the Q-function. The optimal policy can then be derived by a one-step search.

\begin{algorithm}
    \caption{Q-learning}
    \label{qlearning}
    \begin{algorithmic}[1] 
        \Procedure{Q-learning}{$Q$} 
            \Repeat \Until{terminal state}
            \For{each episode}
            	\State Choose action $a$ from policy derived from Q ($\epsilon$ -greedy)
            	\State Take action $a$, observer reward $R$, and next state $S'$
            	\State $ Q(S,A) \gets Q(S,A) + \alpha[R + \gamma\max_{a}Q(S',A) - Q(S,A)]$
            	\State $S \gets S'$
            \EndFor
        \EndProcedure
    \end{algorithmic}
\end{algorithm}

Algorithm \ref{qlearning} shows the pseudocode for Q-learning. The algorithm is initialized with an initial estimate of the Q function. At each step, the agent takes an action $a$ that is derived from the current $Q$ value (Step 5), and observes the reward $R$ (Step 6). The derivation of an action from a given $Q$ function is based on the equation: $\argmax_{a}Q(S,A)$. The essence of $Q$-learning is step 7, which incrementally updates the older $Q$ function estimate based on the reward received in step 6. Difference between the new and old estimates of $V$ or $Q$ function, in this form is commonly referred to as the TD-error.

\subsubsection*{Direct policy search - REINFORCE}

REINFORCE is direct policy search algorithm which maintains an approximation of the policy function $\pi(a|s)$. The actual policy function is approximated with a parameterized representation of the form $\pi(a|s,\theta)$, where $\theta$ is the parameters of the function. Algorithm \ref{reinforce} presents the psudocode of the algorithm.

In every iteration, similar to $Q$-learning, the algorithm updates the policy through stochastic gradient ascent on the parameter $\theta$, thus improving the estimate of the policy representation $\pi(a|s,\theta)$.   

\begin{algorithm}
    \caption{REINFORCE}
    \label{reinforce}
    \begin{algorithmic}[1] 
        \Procedure{REINFORCE}{$\theta$} 
            \State repeat:
            	\State Generate an episode $S_0,A_0,R_1,...S_{T-1},A_{T-1},R_T$, following $\pi(.|.,\theta)$
            	\For{each step of the episode}
            	\State $G_t \gets$ return from step $t$
            	\State $ \theta \gets \theta + \alpha \gamma^t G_t \Delta_\theta \log \pi(A_t|S_t,\theta)$
            	\EndFor
        \EndProcedure
    \end{algorithmic}
\end{algorithm}

\subsubsection{Actor-critic algorithm}

Algrithm \ref{actcritic} presents the psudocode of hte algorithm. Actor-critic combines the above algorithms into a single algorithm, where a temporal-difference is calculated to update the value function. This is in principle similar to TD-learning, the only difference being that the value function is represented in a parametric form $v(s,\mathbf{w)}$. Thus, the value function is updated as stochastic gradient update of the weights $\mathbf{w}$ (step 7), instead of a step update like in TD-learning. This \textit{temporal-difference} is also used to update the parameters of the policy function $\pi(a|s,\theta)$ (step 8), similar to REINFORCE. Detailed derivation and description of the notation is provided in the footnote link \footnote{http://ufal.mff.cuni.cz/\textasciitilde straka/courses/npfl114/2016/sutton-bookdraft2016sep.pdf\#page=288}.
\begin{algorithm}
    \caption{Actor Critic algorithm}
    \label{actcritic}
    \begin{algorithmic}[1] 
        \Procedure{Actor-critic}{$\mathbf{w},\theta$} 
            \State Repeat forever:
            \While{terminal state not reached}
            	\State Choose action $a$ according to the actors current policy $\pi(.|S,\theta)$
            	\State Take action $a$, observer reward $R$, and next state $S'$
            	\State $ \delta \gets R + \gamma \hat{v}(S',\mathbf{w}) - \hat{v}(S,\mathbf{w})$
            	\Comment{temporal difference error}
            	\State $\mathbf{w} \gets \mathbf{w} + \beta \delta \Delta_{\mathbf{w}} \hat{v(S,\mathbf{w})}$
            	\Comment{gradient update of critic's value function parameters}
            	\State $\theta \gets \theta + \alpha I \delta \Delta_\theta \log \pi(A|S,\theta)$
            	\Comment{gradient update of actors's policy parameters}
            	\State $I \gets \gamma I$
            	\State $ S \gets S'$
            \EndWhile
        \EndProcedure
    \end{algorithmic}
\end{algorithm}

\begin{figure}[htbp]
\center{
  \includegraphics[width=0.4\linewidth]{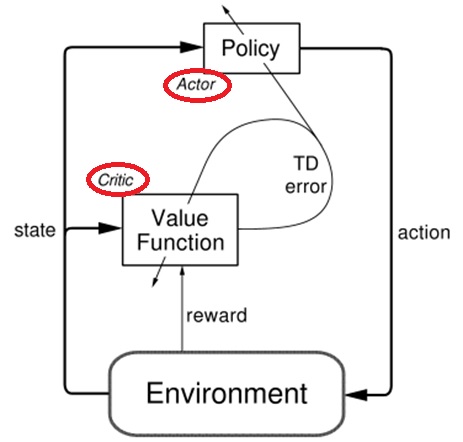}
  \caption{Schematic representation of actor-critic algorithm.}
  \label{fig:actor-critic}
  }
\end{figure}

\subsubsection*{A3C: Asynchronous Advantage Actor-Critic algorithm}

\begin{figure}[htbp]
\center{
  \includegraphics[width=0.7\linewidth]{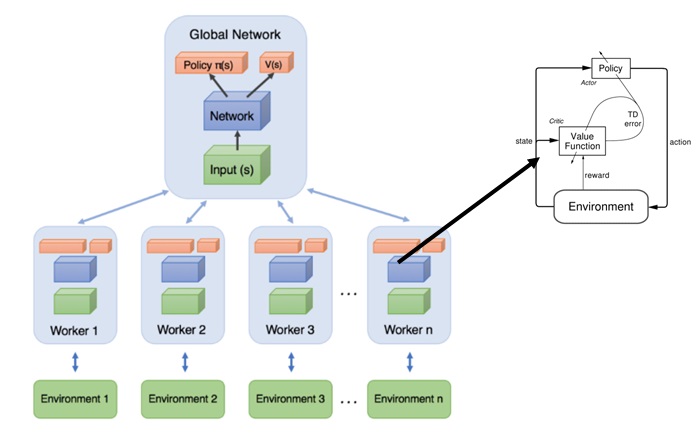}
  \caption{Schematic representation of Asynchronous Advantage Actor Critic algorithm (A3C) algorithm.}
  \label{fig:a3c}
  }
\end{figure}

Asynchronous Advantage Actor Critic (AC3) algorithm is a parallel asynchronous multi-threaded implementation of actor-critic algorithm (Figure \ref{fig:a3c}) \cite{mnih2016asynchronous}. The psudocode of A3C is shown in algorithm \ref{algo:a3c}. Each thread is identical to actor-critic algorithm, however, instead of updating gradients at each step, like in actor-critic, the A3C algorithm accumulates the gradients (steps 6,7), and performs an asynchronous update to the global actor-critic unit after $T_{MAX}$ steps (steps 9,10).

\begin{algorithm}
    \caption{A3C pseudocode for each thread}
    \label{algo:a3c}
    \begin{algorithmic}[1] 
        \Procedure{A3C}{$\mathbf{w},\theta$} 
            \State repeat:
            	\State Generate an episode $S_0,A_0,R_1,...S_{{T_{MAX}}-1},A_{{T_{MAX}}-1},R_{T_{MAX}}$, following $\pi(\cdot|\cdot,\mathbf{\theta})$
            	\For{each step $i$ of the episode}
            	\State $ \delta \gets R_i + \gamma \hat{v}(S',\mathbf{w}) - \hat{v}(S,\mathbf{w})$
            	\Comment{temporal difference error}
            	\State $d\mathbf{w} \gets d\mathbf{w} + \beta \delta \Delta_{\mathbf{w}} \hat{v(S,\mathbf{w})}$
            	\Comment{gradient accumulation of critic's value function parameters}
            	\State $d\theta \gets d\theta + \alpha I \delta \Delta_\theta \log \pi(A|S,\theta)$
            	\Comment{gradient accumulator of actors's policy parameters}
            	\EndFor
	           	\State Perform asynchronous update of global $\mathbf{w}$ using $d\mathbf{w}$ for global critic
	           	\State Perform asynchronous update of global $\theta$ using $d\theta$ for global actor
        \EndProcedure
    \end{algorithmic}
\end{algorithm}

\section{Philosophy of Robert Brandom}

\subsection*{Background}

Robert Brandom (b.1950) is a Distinguished Professor of Philosophy at the University of Pittsburgh, where he has been a member of the faculty since 1976. He completed his PhD at Princeton University under the guidance of Richard Rorty. Most notable of Brandom's work include \textit{Making It Explicit}, and a set of lecture publications delivered at the University of Oxford titled \textit{Between Saying and Doing}.
Brandom's philosophy is grounded on the idea of pragmatism \footnote{https://plato.stanford.edu/entries/pragmatism/}, and his work falls under the philosophy of language, philosophy of mind, and philosophy of logic. His main contribution is inferentialism, where he shows how the process of making inferences imparts semantic meaning to things \cite{brandom1994making}\cite{brandom2008between}.

\subsection{Sapience}
\label{sec:sapience}

The starting point of Brandom's philosophy is the question:
\begin{displayquote}
What is the difference between a parrot who is disposed reliably to respond differentially to the presence of red things by saying "Raawk, that's red". and a human reporter who makes the same noise under the same circumstances \cite{brandom1995knowledge}.
\end{displayquote}

Simply put, this question asks \textit{what does it mean to be us?} Asking this question has significant value in the philosophy of AI, where recent advances and widespread acceptance of artificial intelligence technologies have drawn society to introspect on the question what it \textit{means} to be human. Brandom's response to this question is by arguing that a parrot's behavior is not a part of a special kind of norm governed social practice. Building up on the same argument, he presents a layer-cake picture of sapience, leading him to list \textit{three} types of beings\footnote{The terminology of \textit{logical} and \textit{rational} is of Brandom's own choosing, and might not translate directly to a colloquial understanding of these words.}, ordered based on their sapience \cite{wanderer2014robert}. 

\textit{Simple performers:} Simple performers, or the type Brandom calls Reliable Differential Responsive Dispostion, are reactive systems. They can be imagined to have a read and write head, where based on an input signal, it has the ability to differentially produce certain types of response outputs.

\textit{Rational beings:} One step higher in the chain of sapience, are \textit{rational beings}. Brandom says that what differentiates rational beings from simple performers is their ability to make moves in a Game Of Giving and Asking for Reasons (GOGAR). Brandom presents a detailed elaboration of what this game constitutes, and we would look into this in subsequent sections.

Brandom further claims that it is possible for a simple performer to achieve the ability of rational beings, through a more complex deployment of its abilities - the ability to draw \textit{inferences}. For example, if someone is presented with an offer of employment, they have the option of signing it and taking the job. To take this all important decision, a rational being might consider all the consequences of signing that offer, for example, having to wake up 6 a.m. every weekday, being able to earn money. Thus, this interaction of signing the offer might be imagined to be an \textit{input-output} relation where the input is the presentation of the offer, and the output is signing it. Although both simple performer and rational beings can take part in this interaction with the help of their abilities, what sets rational beings apart is the ability to draw \textit{inferences} as a consequence of the output action.

\textit{Logical Beings:} Further in the chain of sapience are \textit{logical beings}. They distinguish themselves from rational beings by being able to deploy a special kind of vocabulary, which Brandom terms \textit{elaboration} and \textit{explication}.

\textit{Elaboration} is a relation between two practical abilities. For instance the relation between the ability to do multiplication and subtraction (P1), and the ability to do long-form division (P2) \cite{wanderer2014robert}. The relation between P1 and P2 is such that, it is possible to achieve the ability P2 entirely from the ability P1 through a step wise process. Elaboration is this process of step-by-step algorithmic derivation to achieve one ability from the other. Other than algorithmic derivation, elaboration also includes another form of derivation i.e. \textit{elaboration-by-training}. This can be understood as the derivation of one ability from another, where the process of derivation is more complex than just a step-by-step algorithmic process. For example, ability to draw a passable picture of a human from the ability to draw a passable picture of a stick figure.

\textit{Explication} is the act of making explicit the principles that codifies a practical know-how. For example, expressing in some linguistic form the ability to ride a bicycle, or swinging a baseball bat is an act of explication. 

Elaboration and explication in conjunction is referred to as LX-vocabulary. This vocabulary can contain normative expressions like '...is committed to...', ascriptional phrases like '...says', or conditional phrases of the form 'if....else...'. Brandom states that the list of LX-vocabulary is unbounded, and any vocabulary that assists in the process of elaboration and explication can be considered a logical vocabulary.

\begin{figure}[htbp]
\center{
  \includegraphics[width=0.7\linewidth]{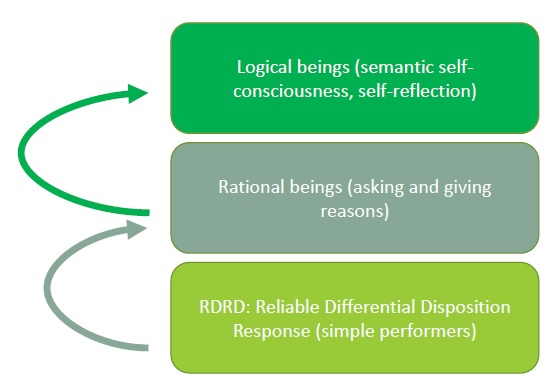}
  \caption{Layer-cake picture of sapience in an increasing order from bottom (simple performers) to top (logical beings)}
  \label{fig:layercake}
  }
\end{figure}

Figure \ref{fig:layercake} shows the three types of beings in increasing order of sapience (bottom to top). Critics of Brandom argue if it is at all possible to conceptualize rational beings as separate from logical beings. My understanding is that rational beings is a stepping stone towards logical beings from simple performers. In relation to Searle's Chinese room thought experiment \cite{hauser2001chinese} that passes a Turing test (or what he calls weak AI), Brandom's rational beings would be an instantiation of weak-AI that passes that Turing test. Thus, whereas a rational being perform a series of statements and inferences, thereby passing a Turing test, it is only logical beings who have the ability to \textit{understand} that performance (strong-AI).

\subsection{Game of Giving and Asking For Reasons (GOGAR)}

\begin{figure}[htbp]
\center{
  \includegraphics[width=0.7\linewidth]{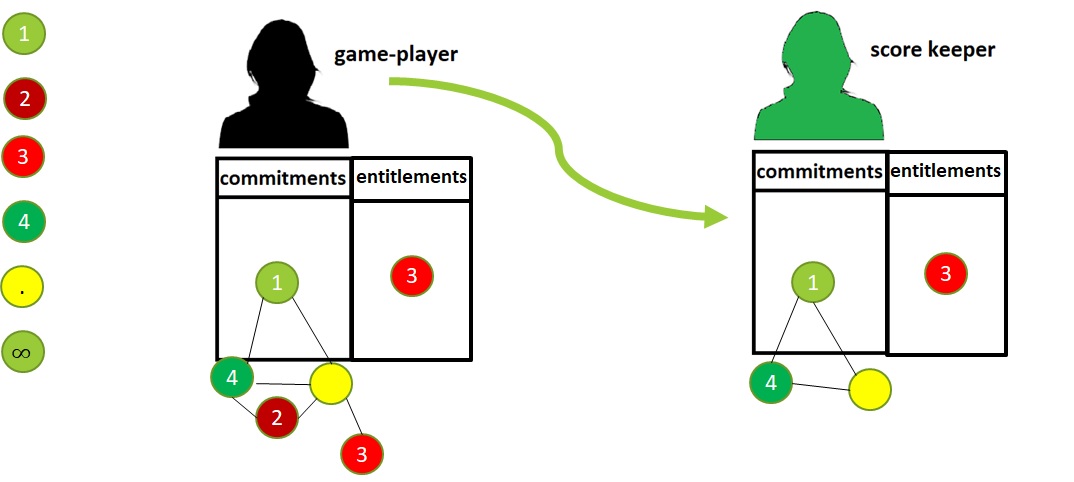}
  \caption{A representation of the Game Of Giving and Asking for Reasons}
  \label{fig:gogar}
  }
\end{figure}

An essential aspect of Brandom's philosophy is his account of sapience. Whereas the previous section presented his work on the three categories and their hierarchy in the ladder of sapience, this section presents his account of the necessary and sufficient conditions that can qualify a sapient being (rational or logical). This can be understood through the interaction of a being in a given social practice\footnote{A social practice can be imagined to be any interaction among beings (humans or not). E.g. in relation to our previous example, engaging in the process of employment can be an example of a social practice.}, or what \textit{form} those interactions should be, to qualify as sapient (rational or logical). Brandom does this by mapping a being's interaction within a social practice to a specific game - the Game of Giving and Asking For Reasons (GOGAR) (Figure \ref{fig:gogar}). 

The main elements of the GOGAR game are \footnote{For the sake of purity, I have retained the original terminologies of the game without modification. These terminologies sometimes seem unintuitive. Thus, whenever possible, I have tried to add subjective descriptions for clarity.}
\begin{itemize}
\item \textit{Players} and \textit{Scorekeepers}: Each individual in the game acts as players (while making a move), as well as scorekeepers to other player's moves.
\item \textit{counters:} There are infinite distinct counters in the game. The counters ($C$) can be thought to be tokens. Each counter is related to other counters through a relation of committive consequences (cc):
\[
C_0 \xRightarrow{\text{cc}} (C_1 \wedge C_2 \wedge C_3 \wedge ...)
\]

\item \textit{commitment move:} This is an act of the player taking a counter and placing it in their commitment box. As a part of the move, the player also has to place all other counters that participate in the committive consequence relation to the chosen counter. This can be interpreted as an act of making a claim (either through statements or through actions). Forcing the player to place the counters in $cc$ relation can be interpreted as being held to the consequences of making that claim. For example making a claim of 'signing an employment offer letter' might draw committive consequence of 'coming to work at 9am' and 'wearing a bowtie'.

\item \textit{entitlement move:} Similar to commitment move, players can place copies of the counters already in their commitment boxes into entitlement boxes. The difference between entitlement and commitment is that a player is obligated to defend an entitlement, if it is challenged by a scorekeeper. A challenge by the scorekeeper can be understood as the scorekeeper not accepting a claim. For example, if the employee claims 'receiving travel allowance' as a $cc$ of 'signing an employment offer letter', then the HR manager (acting as a scorekeeper) can choose not to accept that claim. The player would have to then remove that claim from their entitlement box.

\item \textit{A-ing (asserting):} A scorekeeper can consider a claim by the player as an a-ing move iff the claim is an entitled one \textit{and} the player is committed to defend that entitlement if challenged by any other scorekeeper.
\end{itemize}

\section{Reinforcement learning semantics and Brandom-ian philosophy}

As we discussed in the previous section, both commitments and entitlements use the relation of \textit{commitive consequnce} as a mode to express claims. Even the scorekeeper's challenge to players are expressed through the modality placing counters (and counters that form a $cc$ relation to that counter), that are in opposition to the counters placed by the player. Thus, the counters and the relation of $cc$ is the basic structure on which the GOGAR game is built. It is also relevant here to refresh the intuitive explanation of the counter and the \textit{act} of placing the counter. Whereas a counters represents the set of all possible claims, the act of placing a counter is the pragmatic action of \textit{execution} of that claim (through saying or doing something) by the player.

Thus, one way we relate reinforcement learning semantics to Brandomian philosophy is by connecting the structural elements of RL, to the structural elements of GOGAR.

\begin{figure}
\begin{center}
\begin{subfigure}{.7\textwidth}
  \centering
  \includegraphics[width=.8\linewidth]{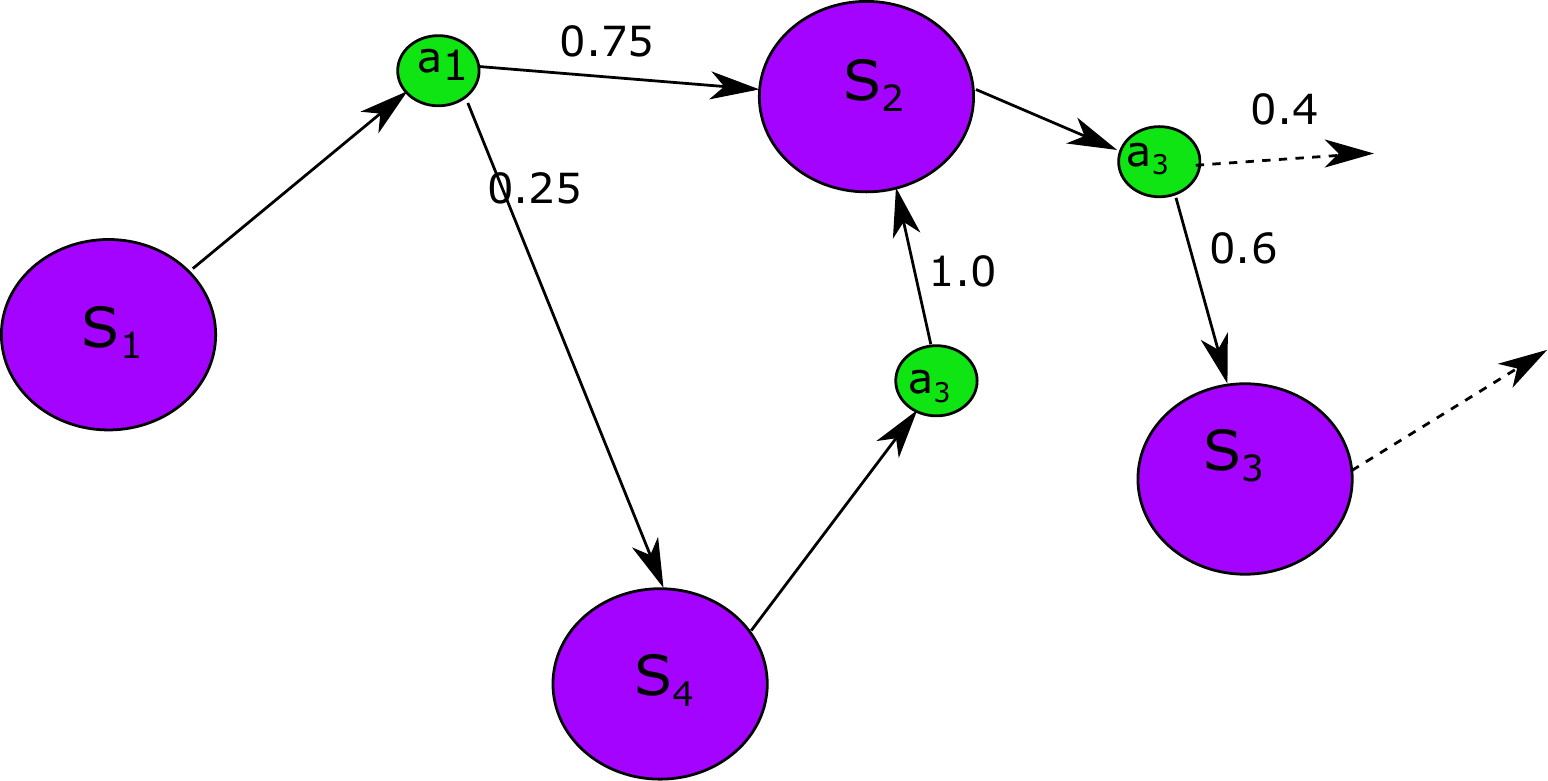}
  \caption{Original policy representation}
  \label{fig:sfig1}
\end{subfigure}\
\begin{subfigure}{.7\textwidth}
  \centering
  \includegraphics[width=.8\linewidth]{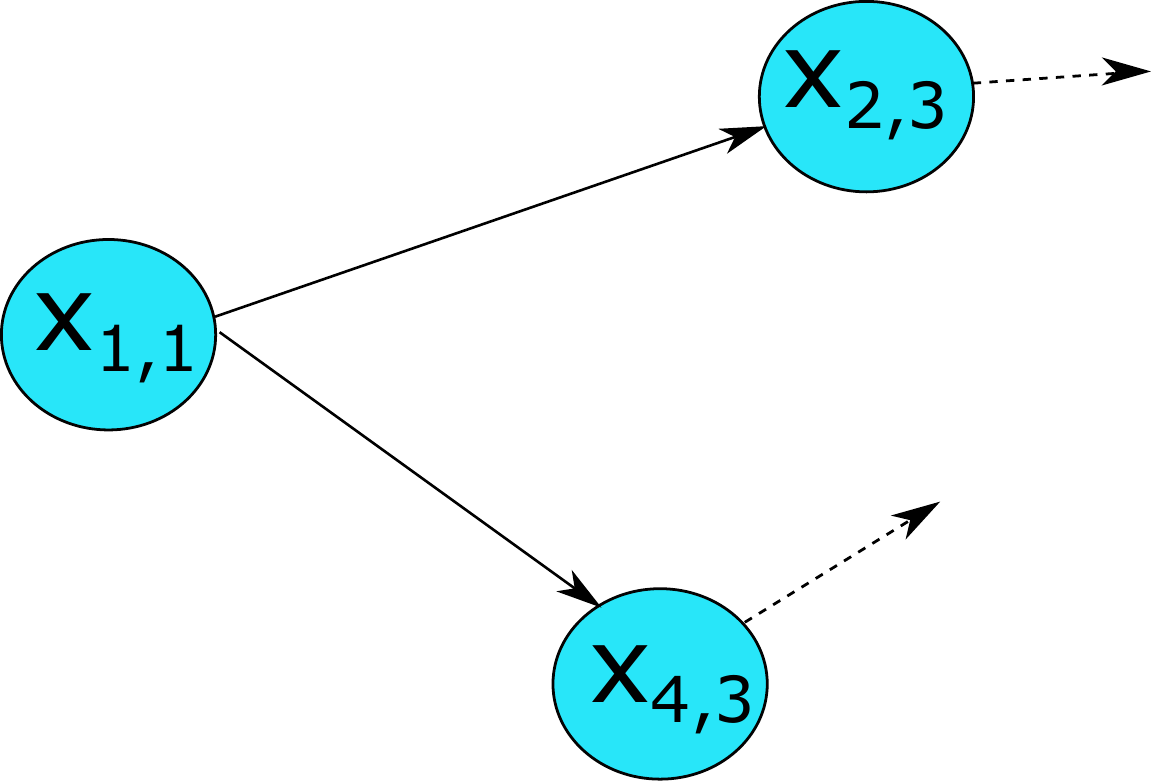}
  \caption{Restructured policy representation}
  \label{fig:sfig2}
\end{subfigure}
\caption{Policy function representation in an MDP}
\label{fig:mdp_infer}
\end{center}
\end{figure}

An MDP distinguishes between a \textit{state} ($S$) and an \textit{action} $A$, whereas GOGAR does not have such explicit distinction. It uses counters and the committive consequence relation among them. Reproducing our earlier notation, this structure in GOGAR can be represented as:
\begin{equation}
C_0 \xRightarrow{\text{cc}} (C_1 \wedge C_2 \wedge C_3 \wedge ...)
\end{equation}
In reinforcement learning, a policy function maps a state to an action \footnote{A stochastic policy function is a probability distribution on the states and the action $\pi(s,a)$. However, a deterministic policy is a special case where, given a state, $\pi(s,a)$ is 1 for at most one action, and 0 for all other. Although we consider only deterministic policies here; extension to stochastic policies are left as a future work}, which can be represented in the form (refer Figure \ref{fig:mdp_infer}):

\begin{equation}
s_1 \rightarrow a_1 \wedge s_2 \rightarrow a_3 \wedge ...
\end{equation}

If we combine states and actions into a tuple such that
\begin{equation}
x_{i,j}=(s_{i},a_{j}) \in X | \pi(s_{i},a_{j}) =1 \quad\forall i,j \in \mathbb{N}\\
\end{equation}
\begin{equation}
x_{i,j} \rightarrow x_{m,n}= 
\begin{cases}
    true,& \text{if } T(s_{i},s_{m})\geq 0\\
    false,              & T(s_{i},s_{m})= 0
\end{cases}
\end{equation}

where $T(s,a)$ is the transition function, then the same policy function can be represented as
\begin{equation}
x_{1,1} \rightarrow (x_{2,3} \wedge x_{4,3}) ...
\end{equation}

Thus, with the above reformulation, we can see that formulation (1) of GOGAR and formulation (5) of reinforcement learning are structurally equivalent. This brings us to our link 1:
\begin{displayquote}
\textit{Link 1:} \textit{The commitive consequence relation among tokens in GOGAR is structurally equivalent to a policy in reinfocement learning.}
\end{displayquote}

Our treatment of the above link between RL and Brandom's philosphy also gives us a fresh perspective to reinforcement learning and MDP. We often view an MDP as a barebone mathematical model devoid of any innate meaning. It gets it's meaning only through how an MDP and reinforcement learning is applied to a specific application. However, our conjecture is this need not be the case. Since the state-action tuple ($x_{i,j}$) is equivalent to Brandom's use of counters ($C$) in GOGAR, we can use state-action tuples ($x_{i,j}$), as basic building blocks towards a strong-AI, just like how the counters build up towards the sapience of logical beings \footnote{This is directly derived by drawing parallels between Searle's use of strong-AI, and Brandom's \textit{logical beings} being an instance of that. }

Next, we delve into the \textit{semantics} of the counters and committive consequence relation in GOGAR. Brandom's position on semantics of the committive consequence is related to a \textit{constitutive} view of a social practice. In other words, the specific social practice in which the counters and $cc$ relations appear, is the source of the semantics. Intuitively, it helps to understand this in reference to our previous example. The commitive consequence of 'signing an employment offer' to 'waking up at 6am' has meaning \textit{only because} of the social practice of 'engaging in the process of employment'. Thus, the social practice have a constitutive function in relation to committive relationships \cite{evans2016computer}.

More formally, this means that any committive consequence in the formulation (1) needs to be indexed in reference to the social practice $p \in \mathbb{P}$, where $\mathbb{P}$ is the superset of all social practices.
\begin{equation}
[C_0 \xRightarrow{\text{cc}} (C_1 \wedge C_2 \wedge C_3 \wedge ...) ]^{p} \quad  \mathtt{because} \quad \exists p \in \mathbb{P} 
\end{equation}

Taking a similar line of argument of reinforcement learning, we can claim that RL policy formulation of the form (5), has semantic meaning only through the transition relation $T$. We have already established this structurally in equation (4). However, semantically we can say that

\begin{equation}
[x_{1,1} \rightarrow (x_{2,3} \wedge x_{4,3}) ... ]^{T} \quad  \mathtt{because} \quad \exists T \in \mathbb{T} 
\end{equation}

where $\mathbb{T}$ is the superset of all possible transition functions. This equivalence between equation (6) and (7) brings us to the second link:

\begin{displayquote}
\textit{Link 2:} \textit{The transition function, or the underlying model of the environment in reinforcement learning can be understood as the social practice in a GOGAR.}
\end{displayquote}

\subsection*{Re-imagined actor-critic and A3C}

Having drawn structural and semantic links between reinforcement learning and Brandom's philosophy, we now connect the two on the basis of the reinforcement learning algorithms discussed in section \ref{sec:algos}. We do this by asking the following questions:

\begin{itemize}
\item \textit{1. If being able to play the GOGAR is a necessary condition of sapience, can a reinforcement learning algorithm play the game?}
\item \textit{2. If the path towards a logical being, and strong-AI, is through the deployment of a set of logical vocabulary, can that algorithm deploy that vocabulary?}
\end{itemize}

To address the first question, we restructure the A3C algorithm, already introduced in section \ref{sec:algos}, in line with the GOGAR game, thereby showing that threads participating in AC3 can play the game. \footnote{We consider the case where all the participants in the game (players and scorekeepers) are reinforcement learning based agents. Extension to cases where a portion of the population is non-AI based, is left as a future work}

\begin{figure}[htbp]
\center{
  \includegraphics[width=0.45\linewidth]{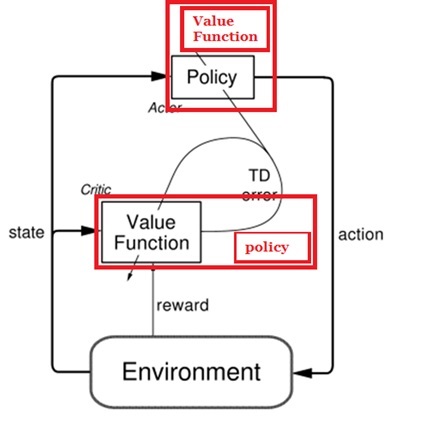}
  \caption{Restructured actor-critic unit in A3C}
  \label{fig:ac-restruct}
  }
\end{figure}

First, we restructure the basic actor-critic units of the A3C. Figure \ref{fig:ac-restruct} shows the new restructured actor-critic unit in an A3C algorithm. Instead of a static assignment of actor and critic, as in the original algorithm, we introduce the idea of a participant unit (\textit{pu}) (shown within the red box in the figure), and actor and critic just being dynamic roles played by a $pu$. Each participant unit holds a value function representation $v(s,\mathbf{w})$, which helps them play the role of a critic, as well as a policy representation $\pi(a|s,\theta)$, which helps the same participating unit play the role of an actor. This is in accordance with the GOGAR where a participant plays the dual roles of player and scorekeeper.

\begin{algorithm}
    \caption{GOGAR A3C}
    \label{gogar-a3c}
    \begin{algorithmic}[1] 
        \Procedure{GOGAR-A3C}{$P$}
        	\State sample $n\_interac$ from [0,$\comb{P}{2}$]
        	\State initialize $n\_interac$ threads 
            \For{each thread}
            	\State $pu\_act \gets$ sample from $P$
            	\State $pu\_crit \gets$ sample from $P \setminus (pu-act)$
            	\While{$t \leq T_{MAX}$}
            	\State Choose action $a$ according to the actors current policy $\pi_{pu\_act}(.|S,\theta)$
            	\State Take action $a$, observer reward $R$, and next state $S'$
            	\State $ \delta \gets R + \gamma \hat{v}_{pu\_crit}(S',\mathbf{w}) - \hat{v}_{pu\_crit}(S,\mathbf{w})$
            	\Comment{temporal difference error}
            	\State $\mathbf{w} \gets \mathbf{w} + \beta \delta \Delta_{\mathbf{w}} \hat{v}_{pu\_crit}(S,\mathbf{w})$
            	\Comment{gradient update of critic's value function parameters}
            	\State $\theta \gets \theta + \alpha I \delta \Delta_\theta \log \pi_{pu\_act}(A|S,\theta)$
            	\Comment{gradient update of actors's policy parameters}
            	\State $I \gets \gamma I$
            	\State $ S \gets S'$
            	\State $t \gets t+1$
            	\EndWhile
            \EndFor
        \EndProcedure
    \end{algorithmic}
\end{algorithm}

\begin{figure}[htbp]
\center{
  \includegraphics[width=0.6\linewidth]{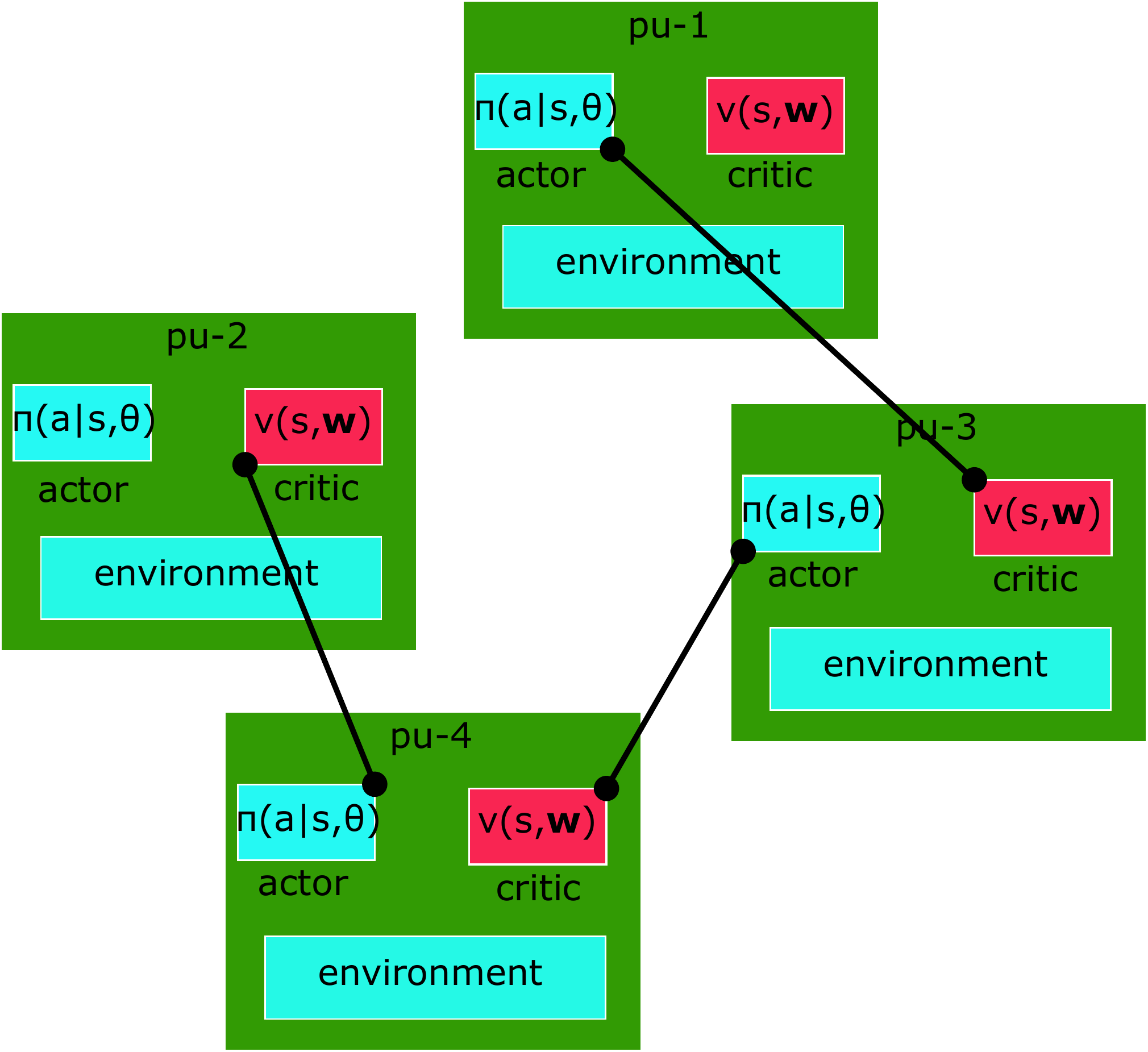}
  \caption{Restructured AC3 with 4 participating units and 3 interactions}
  \label{fig:ac3-restruct}
  }
\end{figure}

Next, we restructure the complete algorithm as shown in Figure \ref{fig:ac3-restruct}. Algorithm \ref{gogar-a3c} shows the psudocode of the algorithm. The algorithm is initialized with a population ($P$) of participating units. Next, we initialize $n\_interac$ threads, each simulating an interaction between two participating units, each sampled without replacement from the population (steps 5,6). Until a predefined length of an interaction ($T_{MAX}$), each thread then executes an actor-critic algorithm (steps 8-14), with the assigned actor and the critic. Further elements of the game like commitments and entitlements can then be designed as primitives, depending on the specific application.

Having addressed the first question by restructuring the A3C algorithm so that the agents can participate in the game of GOGAR, we now focus on the second question. 

As mentioned in section \ref{sec:sapience}, participants in the game of GOGAR can advance themselves from rational beings to logical beings by deploying \textit{logical vocabulary}. One category of logical vocabulary is the conditional locution of \textit{if...then}. These condition locutions can also be interpreted as prediction questions of the form \textit{\textbf{if} I do this....\textbf{then} does that happen?} In our restructured implementation of GOGAR-AC3, each participating units hold a value function representation. These value functions, in principle, can also be substituted by \textit{general value functions (GVF)}. GVFs are a generalization of standard value function in reinforcement learning, such that agents are able to answer prediction questions of the aforementioned form \cite{sutton2011horde}. Although, a step-by-step re-implementation of GOGAR-AC3 with \textit{GVFs} instead of \textit{value-functions} are left as a future work, with reference to our second question, we feel confident to answer that it is in principle possible to deploy logical vocabulary in GOGAR through reinforcement learning.
 
\section*{Conclusion}

In this paper, we presented a Brandom-ian view of reinforcement learning with the objective of advancing it towards strong-AI. We introduced the main elements of reinforcement learning, MDPs, and presented some popular RL algorithms. Next, we presented  Brandom's philosophy, and described in detail, the mechanism of the Game Of Giving and Asking For Reasons (GOGAR). As a main contribution, we drew links between the constituting elements of reinforcement learning and GOGAR. Further we theorize on two important questions, that we believe lie in the path towards strong-AI. Finally, we show that it is possible to re-imagine reinforcement learning, with the help of Brandom's philosophy, for the objective of achieving strong-AI.


\bibliographystyle{unsrt}

\bibliography{sample}


\end{document}